\documentclass[10pt,twocolumn,letterpaper]{article}
\usepackage[final]{cvpr}

\usepackage[utf8]{inputenc} 
\usepackage{url}            
\usepackage{amsfonts}       
\usepackage{nicefrac}       
\usepackage{graphicx}
\usepackage{wrapfig}

\usepackage{graphicx}
\usepackage{amsmath}
\usepackage{amssymb}
\usepackage{booktabs}
\usepackage{placeins}
\usepackage{tabularx}
\usepackage{multirow}
\usepackage{makecell}

\usepackage[dvipsnames]{xcolor}

\definecolor{cvprblue}{rgb}{0.21,0.49,0.74}
\usepackage[pagebackref,breaklinks,colorlinks,citecolor=cvprblue]{hyperref}

\def\paperID{20} 
\def\confName{CVPR}
\def\confYear{2024}

\def\confName{ar$\xi$iv}
\def\confYear{2024}

\title{A Method of Moments Embedding Constraint and its Application to Semi-Supervised Learning}

\author{Michael Majurski\\
	Information Technology Lab, NIST\\
	University of Maryland, Baltimore County\\
	{\tt\small michael.majurski@nist.gov}
\and
Sumeet Menon\\
University of Maryland, Baltimore County\\
{\tt\small sumeet1@umbc.edu}
\and
Parniyan Farvardin\\
University of Miami\\
{\tt\small pxf291@miami.edu}
\and
David Chapman\\
University of Miami\\
{\tt\small dchapman@cs.miami.edu}
}

\begin{document}
\maketitle

\begin{abstract}
	Discriminative deep learning models with a linear+softmax final layer have a problem: the latent space only predicts the conditional probabilities $p(Y|X)$ but not the full joint distribution $p(Y,X)$, which necessitates a generative approach.
	The conditional probability cannot detect outliers, causing outlier sensitivity in softmax networks.
	This exacerbates model over-confidence impacting many problems, such as hallucinations, confounding biases, and dependence on large datasets.
	To address this we introduce a novel embedding constraint based on the Method of Moments (MoM).
	We investigate the use of polynomial moments ranging from 1st through 4th order hyper-covariance matrices.
	Furthermore, we use this embedding constraint to train an Axis-Aligned Gaussian Mixture Model (AAGMM) final layer, which learns not only the conditional, but also the joint distribution of the latent space.
	We apply this method to the domain of semi-supervised image classification by extending FlexMatch with our technique.
	We find our MoM constraint with the AAGMM layer is able to match the reported FlexMatch accuracy, while also modeling the joint distribution, thereby reducing outlier sensitivity.  We also present a preliminary outlier detection strategy based on Mahalanobis distance and discuss future improvements to this strategy. 
	Code is available at: \url{https://github.com/mmajurski/ssl-gmm}
\end{abstract}

\section{Introduction}

\begin{figure}[ht]
	\centering
	\includegraphics[width=0.8\linewidth]{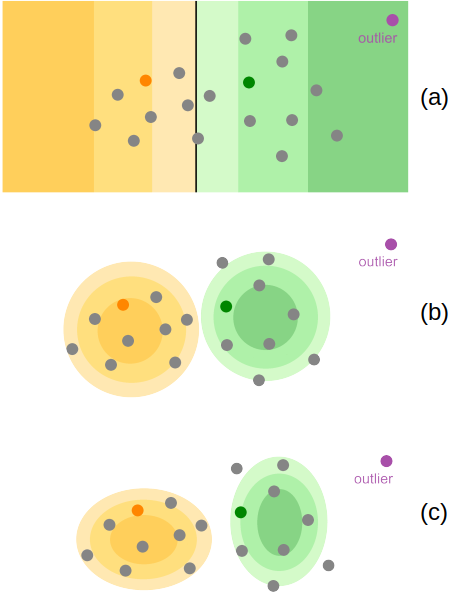}
	\caption{Schematic of the outlier problem, and how generative modeling of the joint probability can improve the situation.  Prediction with (a) fully-supervised softmax, (b) semi-supervised KMeans, and (c) semi-supervised AAGMM.} 
	\label{fig:schema}
\end{figure}

\begin{figure*}[ht]
	\centering
	\begin{subfigure}[t]{.24\textwidth}
		\includegraphics[width=\textwidth]{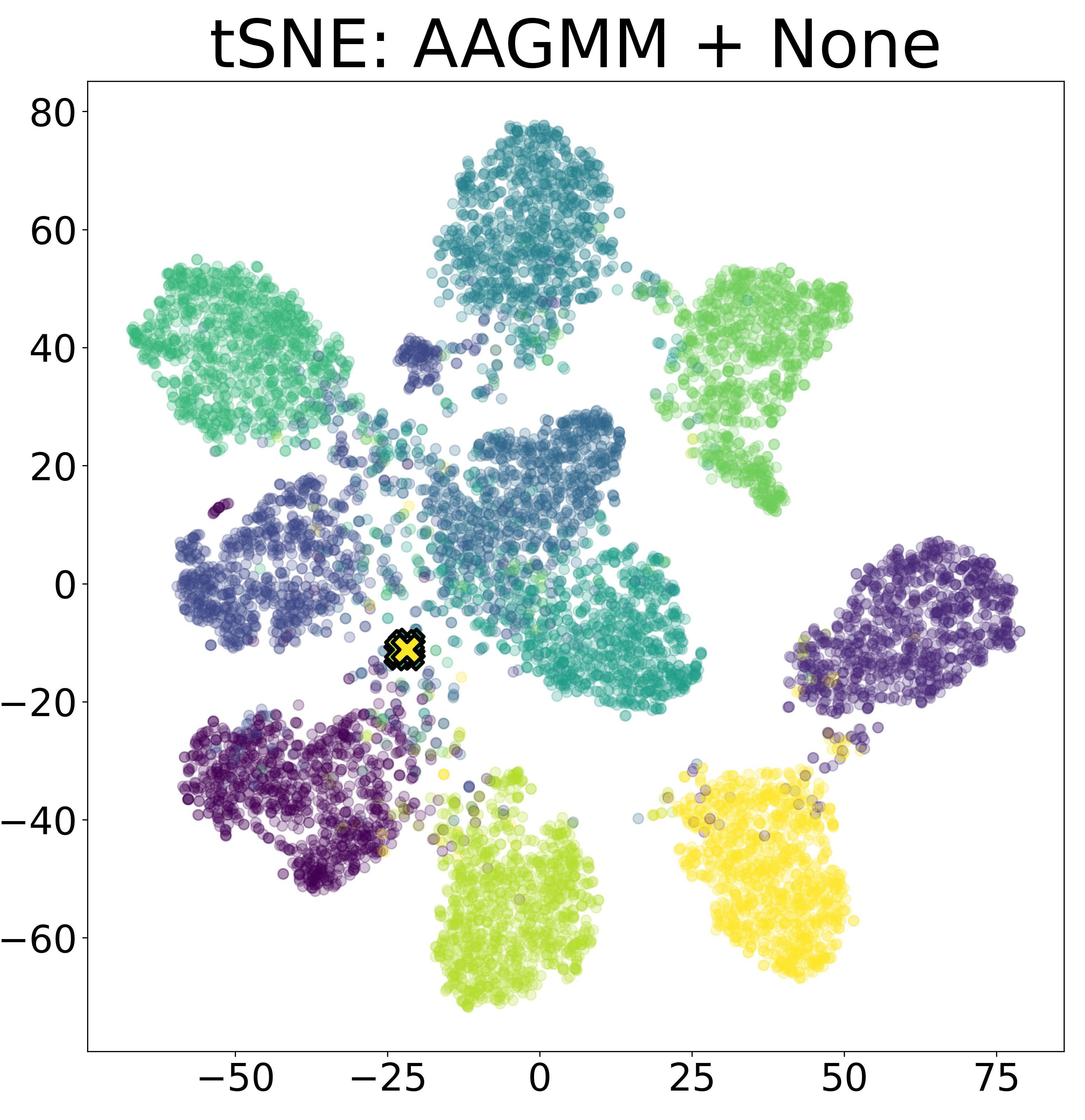}
		\subcaption{}
	\end{subfigure}
	\begin{subfigure}[t]{.24\textwidth}
		\includegraphics[width=\textwidth]{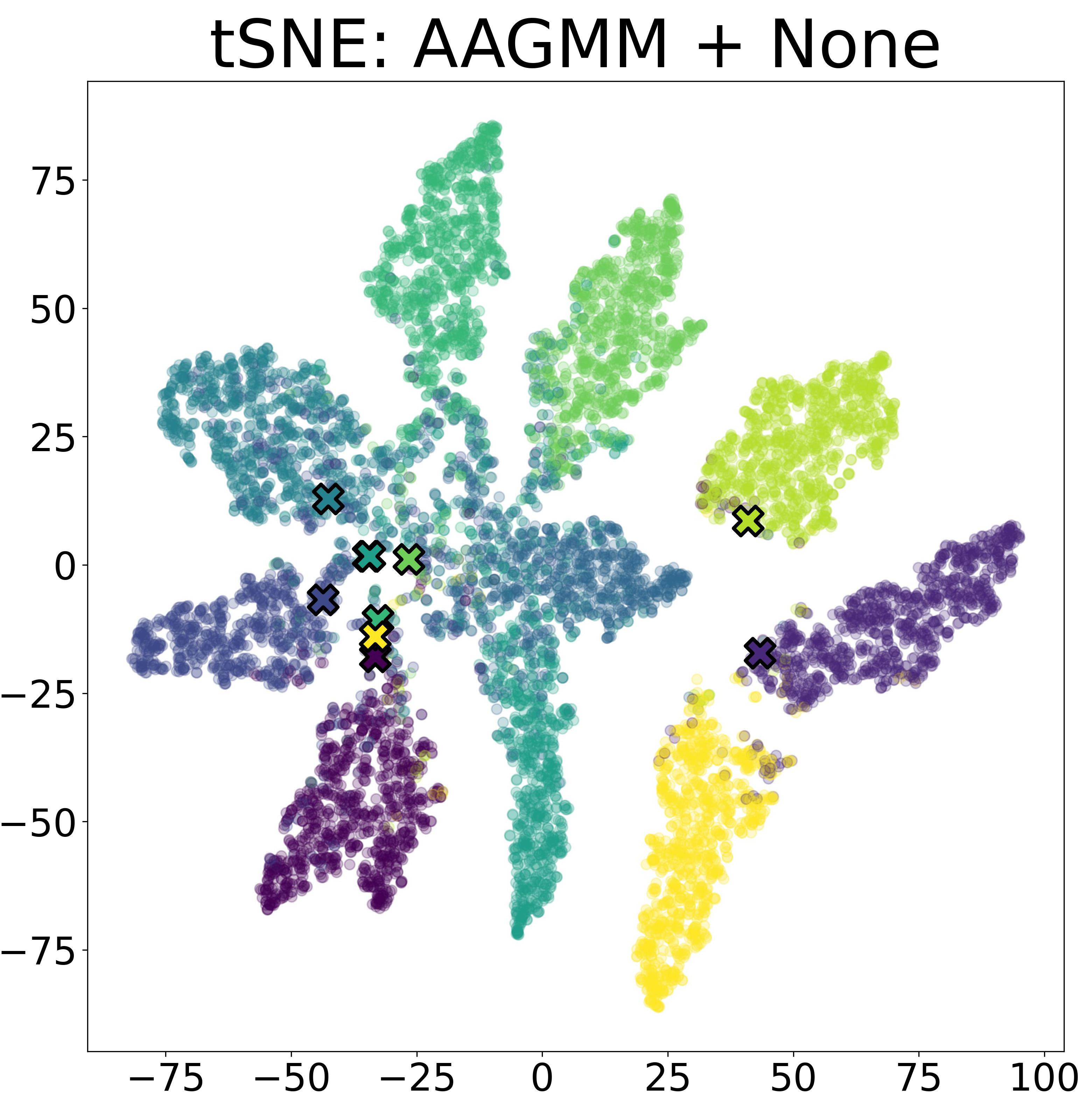}
		\subcaption{}
	\end{subfigure}
	\begin{subfigure}[t]{.24\textwidth}
		\includegraphics[width=\textwidth]{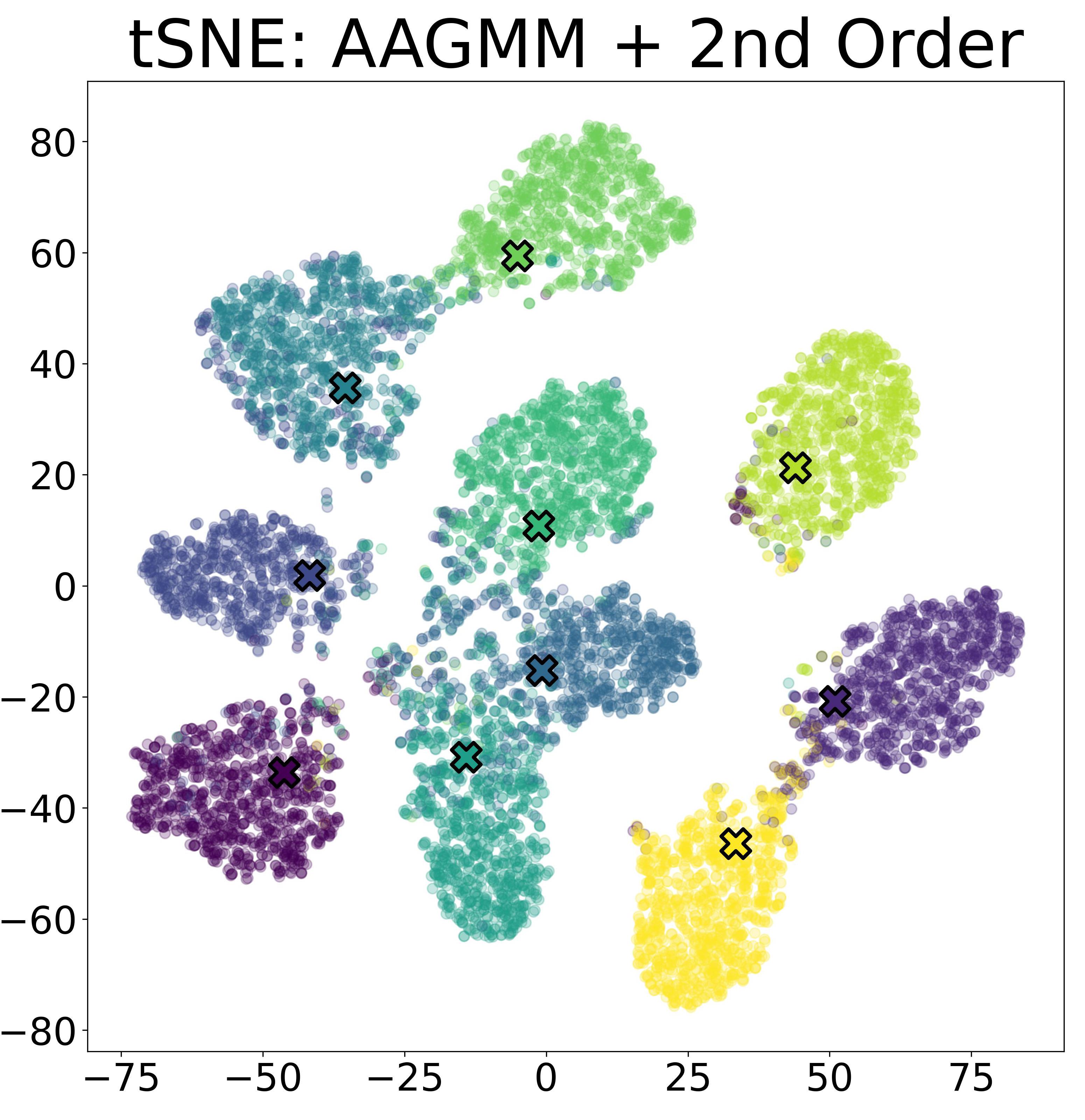}
		\subcaption{}
	\end{subfigure}
	\begin{subfigure}[t]{.24\textwidth}
		\includegraphics[width=\textwidth]{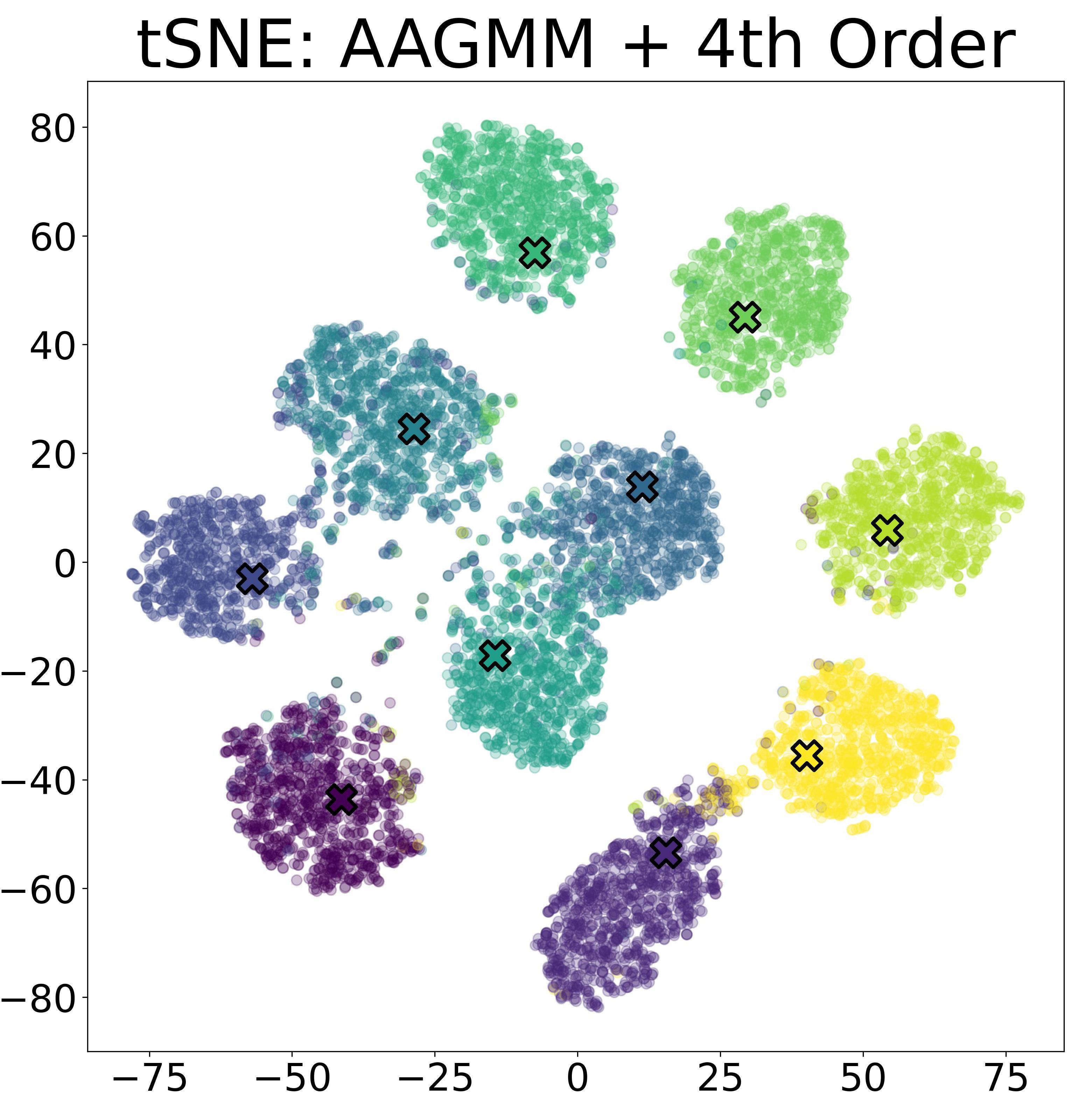}
		\subcaption{}
	\end{subfigure}
	\caption{t-SNE\cite{tsne} plot of the latent embedding space for various final layers with different MoM embedding constraints. AAGMM results include the explicitly modeled cluster centered marked with "X"s. Depending on the run, AAGMM cluster centers can be degenerate (a), non-degenerate but still mis-aligned with the clusters (b), acceptably aligned (c), or well aligned with the underlying clusters (d).} 
	\label{fig:cifar10tsne}
\end{figure*}

The majority of deep classifiers rely on a softmax final activation layer which predicts the conditional probability $p(Y|X)$.
When that layer receives input $X$, the model predicts a soft pseudo-distribution of labels $Y$ which argmax can convert into a hard label.
If $X$ is distant from the decision boundary, then by definition, softmax assigns a prediction $Y$ with high confidence.
This works well for inlier samples, well represented by the training distribution.
However, when presented with an outlier $X$, it is likely $X$ will not be near the decision boundary (Figure \ref{fig:schema}, top).
Therefore, softmax perceptrons, by definition, over-confidently hallucinate when given unexpected inputs \cite{wei2022mitigating}. 
Most deep classifiers use softmax without a safety net and as such over-confidently predict $Y$.
Ideally, when input $X$ is distant from the decision boundary and training exemplars, the model should not be confident about the output class label $Y$.

Replacing softmax with a generative method that models the joint probability $p(Y,X)$ can improve the capability of deep classifiers. 
Models using a final layer capable of learning the joint probability $p(Y,X)$ can infer the conditional $p(Y|X)$.
More importantly, such a layer can also infer the prior probability $p(X)$.
Thus, if $X$ is an unexpected input, then such a layer can flag the input as a low-probability sample (equivalently a high-probability outlier), rather than simply confidently predicting a label.

Prior work has explored generative modeling for image classification \cite{li2019disentangled,kingma2013auto,kingma2019introduction}.
How to best train and utilize generative modeling within a deep learning context remains an open question.
The naive approach of minimizing cross entropy between $Y$ and $\hat{Y}$ will not work.
Figure \ref{fig:cifar10tsne} (a \& b) shows t-SNE plots of the model latent embedding space for $94\%$ accurate semi-supervised CIFAR-10 \cite{cifar10} image classification models, demonstrating why the naive approach will not work as intended. 
However, the explicitly modeled cluster centers (shown as X's) do not align with the underlying data.
While that model has acceptable predictive performance, it does not accurately learn and represent the underlying training data.
To construct a robust model, one cannot simply fit a decision boundary.
The model needs to learn the full joint distribution of the latent space. 
Figure \ref{fig:cifar10tsne} (c \& d) demonstrates that with our proposed AAGMM final layer with either 2nd or 4th order MoM embedding constraints, the exact same model can achieve comparable (if not better) accuracy, but with the added benefit of modeling the underlying data clusters in the latent space.

We apply this work to the domain of Semi-Supervised Learning (SSL), because over-confident label predictions can cause confounding issues with pseudo-labeling methods \cite{arazo2020pseudo}.
SSL leverages an abundance of unlabeled data to improve deep learning based model performance under limited training data regimes \cite{zhu2022introduction,li2019safe,hady2013semi}.
Contrastive learning methods leverage the intuition that similar instances should be close in the representation space, while different instances are farther apart \cite{yang2022class,li2021comatch}.
Consistency regularization borrows the intuition that modified views of the same instance should have similar representations and predictions \cite{sohn2020fixmatch,lee2022contrastive,zhang2021flexmatch,kim2022conmatch}.
This work contributes:

\begin{enumerate}
	\item A novel Method of Moments (MoM) based embedding constraint that enables the model to not only learn the decision boundary but also the latent joint distribution. 
	Moreover, this constraint ensures that each latent cluster exhibits a well-behaved Gaussian shape.
	\item A replacement of the final linear+softmax activation layer of the neural network with either an axis-aligned differentiable Gaussian Mixture Model (AAGMM) or an equal variance version named KMeans trained via back propagation, both of which have explicit modeling of class cluster centroids. 
	\item A preliminary outlier removal strategy based on Mahalanobis distance that is compatible with AAGMM and MoM techniques.
\end{enumerate}

We apply this methodology to the task of semi-supervised image-classification using CIFAR-10 \cite{cifar10} and STL-10 \cite{coates2011analysis} with 40 training labels. 
Only 40 labels were used to provide a sufficiently difficult SSL problem, as SOTA results with higher numbers of labeled samples are approaching fully supervised performance \cite{zheng2023simmatchv2}.
The embedding constraint penalties are applied to all unlabeled data and not just the valid pseudo-labels.  
As such our method fits the latent joint distribution across all of the unlabeled data points, an improvement on baseline pseudo-labeling methods (like FlexMatch \cite{zhang2021flexmatch}) which only fit the conditional distribution to the high confidence pseudo-labels while removing low confidence pseudo-labels.

\section{Related Work}


SSL has shown great progress in learning high quality models, in some cases matching fully supervised performance for a number of benchmarks \cite{zhang2021flexmatch}.
The goal of SSL is to produce a trained model of equivalent accuracy to fully supervised training, with vastly reduced data annotation requirements.
Doing so relies on accurately characterizing inlier vs outlier unlabeled samples.

\subsection{Pseudo-Labeling}
Self-training was among the initial approaches employed in the context of SSL to annotate unlabeled images. 
This involves the initial training of a classifier with a limited set of labeled samples which incorporates pseudo-labels exceeding a predefined threshold into the gradient descent process \cite{yarowsky1995unsupervised, mcclosky2006reranking, olivier2006semi,zhai2019s4l,livieris2019predicting,rosenberg2005semi,menon2020deep}. 
A closely related method to self-training is co-training, where a given dataset is represented as two distinct feature sets \cite{blum1998combining}. 
These independent sample sets are subsequently trained separately using two distinct models, and the sample predictions surpassing predetermined thresholds are utilized in the final model training process \cite{blum1998combining,prakash2014survey}.
A notable approach to pseudo-labeling is the Mean Teacher algorithm \cite{tarvainen2017mean}, which leverages exponential moving averages of model parameters to acquire a notably more stable target prediction. 
This refinement substantially enhances the convergence of the algorithm.

Several papers have attempted to enhance the quality of pseudo-labels to either improve the final model accuracy, improve the rate of convergence, or avoid confirmation bias \cite{arazo2020pseudo}.
Rizve et al. \cite{rizve2021defense} explore how uncertainty aware pseudo-label selection/filtering can be used to reduce the label noise.
Incorrect pseudo-labels can be viewed as a network calibration issue \cite{rizve2021defense} where better network logit calibration might improve results \cite{Xing2020DistanceBased}.
Improvements to the pseudo-labeling process have been demonstrated by imposing curriculum \cite{zhang2021flexmatch} or by including a class-aware contrastive term \cite{yang2022class}.
Leveraging the concept of explicit class cluster centers for conditioning semantic similarity improves final model accuracy \cite{zheng2022simmatch}.
Additionally, improvements have been found in incorporating purely clustering based methods like DINO \cite{caron2021emerging} into semi-supervised methods \cite{fini2023semi}.

\subsection{Consistency Regularization}

Consistency Regularization is a branch of techniques that have been instrumental toward many of the state of the art techniques in semi-supervied learning within the last several years \cite{berthelot2019mixmatch, berthelot2019remixmatch, sohn2020fixmatch, zhang2021flexmatch, li2021comatch, kim2022conmatch, zheng2022simmatch, zheng2023simmatchv2}.  The idea being that augmentation does not typically change the meaning of images.  MixMatch is a semi-supervised pseudolabeling that greatly popularized the use of consistency regularization to ensure that augmentation does not affect the predicted label \cite{berthelot2019mixmatch, berthelot2019remixmatch}.  FixMatch further extended this method by introducing the notion of weak and strong augmentations including the cutout operator to increase the robustness of the regularization \cite{sohn2020fixmatch}.  FlexMatch is a further improvement that introduces a curriculum pseudo-labeling strategy for flexible threshold values \cite{zhang2021flexmatch}.  Co-Match made use of a form of consistency regularization to ensure that strong augmentations shared not only a similar pseudolabel, but furthermore a similar embedding space.  Moreover, a neighborhood graph was constructed for embeddings and pseudolabels and refined via co-learning \cite{li2021comatch}.  Con-match introduced a confidence metric based on the similarity of a basket of augmented embeddings \cite{kim2022conmatch}.  SimMatch introduced a graph-based label propagation algorithm through a low-dimensional latent projection, and utilized multiple forms of consistency regularization including both semantic-level and instance-level consistency terms \cite{zheng2022simmatch, zheng2023simmatchv2}.  

\subsection{Latent Embedding Constraints}

A notable latent embedding constraint that is related, yet substantially different from our approach is the Evidence Lower Bound (ELBO) \cite{kingma2013auto}.
ELBO approximates a latent sample with a variational distribution and constrains the KL-divergence between the variational distribution and a target shape which is typically a multivariate standard normal distribution \cite{kingma2013auto}. 
The main drawback of this approach is that the true KL-divergence is intractable to calculate.  
As such, the posterior must take on a simplified form. 
Most practical implementations use a diagonal posterior which can only penalize simple differences in shape such as mean and standard deviation.  
Arbitrarily complex posteriors are nevertheless possible using the method of Normalizing flows \cite{rezende2015variational,kingma2016improved,caterini2021variational}, which provides an iterative framework based on change of variables although this method is quite involved.  
Our MoM constraint is relatively simple but can also penalize complex differences in shape by constraining 2nd, 3rd, and 4th order hyper-covariance matrices, although we do so by comparing the moments directly.  
This greatly simplifies implementation as we do not need to explicitly construct a posterior distribution.

Another notable embedding constraint is the Maximum Mean Discrepancy (MMD), also known as the two-sample test \cite{gretton2007kernel}.  
MMD was used in the Generative Moment Matching Network \cite{li2015generative} and has since been used extensively for the problem of domain adaptation \cite{WANG2018135,Wang2023rethinking}, in order to constrain the latent projections of the source and target distributions to follow the same distribution.  
MMD is a moment-matching constraint based on the kernel trick and can therefore constrain any difference in shape between two samples including very high order moments. 
Due to the kernel trick requiring proper inner products, MMD can only be used to constrain one sample to another sample.  
It cannot directly constrain sample statistics to population statistics, although it is possible to approximate populations numerically via monte-carlo sampling \cite{zhao2019infovae}.  
Like MMD, our method is based on MoM, but it does not involve the kernel trick, and instead penalizes polynomial moments explicitly thereby enabling the sample embedding to be constrained to an exact target distribution.


\section{Methodology}

In this section, we explore our proposed replacement final activation layers and our embedding space constraints.
Our methodology is based upon the published FlexMatch \cite{zhang2021flexmatch} algorithm as implemented in the USB framework \cite{wang2022usb}, with identical hyper-parameters unless otherwise stated.
We extend FlexMatch with a few minor training algorithm modifications explored in Section \ref{hyperparams}.
FlexMatch \cite{zhang2021flexmatch} is a simple, well performing SSL algorithm.
As such, it serves as a good comparison point for exploring the effect of our contributions.

Both the linear layer replacements and the embedding constraints explored herein represent increasing levels of prescription about how the final latent embedding space should be arranged compared to a traditional linear layer.
The idea of leveraging clusters in embedding space is not new \cite{caron2018deep,caron2020unsupervised,enguehard2019semi}, but we extend the core idea with a novel differentiable model with learned cluster centroids and MoM based constraints.
The MoM constraints do not impose any assumptions outside of applying L2 penalties as described in Section \ref{sec:mom}.

\subsection{Alternate Final Layers}

As we discussed in the introduction traditional final activation layers such as linear+softmax are fully discriminative in that they directly estimate the conditional probability $p(Y|X)$. 
These layers do not estimate $p(X)$ or the joint probabilities $p(Y,X)$.
To overcome this limitation, we present two generative final activation layers: (a) the Axis Aligned GMM (AAGMM) layer and (b) an equal variance version of AAGMM that we henceforth call the KMeans activation layer due to the similarity of the objective function with a gradient based KMeans.

These activation layers are fully differentiable and integrated into the neural network architecture as a module in the same way as a traditional final linear layer. 
As such, they do not require external training and do not use expectation maximization.
They are drop-in replacements for the final linear layer.

Importantly, these activation layers exhibit both discriminative and generative properties. 
The neural network model $F(X;\theta_F)$ transforms the data $X$ into a latent space $Z = F(X;\theta_F)$, and the final activation layer estimates the probability densities $p(X)$, $p(Y;X)$ and $p(Y|X)$ by fitting a parametric model to the latent representation $Z$.

\subsubsection{Axis Aligned Gaussian Mixture Model Layer}

The AAGMM layer defines a set of $K$ trainable clusters, one cluster per label category. 
Each cluster $k=1 \dots K$ has a cluster center $\mu_k$ and cluster covariance $\Sigma_k$. 
The prior probability of any given sample $X_i$ is defined by the mixture of cluster probability densities over the $D$-dimensaional latent representation $Z_i$ as follows,

\begin{equation}
	\begin{aligned}
		\label{eq_px}
		&p(X_i) = \sum_{k=1}^K \mathcal{N} (Z_i, \mu_{k}, \Sigma_k) \\[10pt]
		&\text{where} \quad Z_i = F(X_i, \theta_F)
	\end{aligned}
\end{equation}

Where $\mathcal{N}(Z_i, \mu_k, \Sigma_k)$ represents the multivariate Gaussian pdf with centroid $\mu_k$ and covariance $\Sigma_k$. 
AAGMM is axis aligned because $\Sigma_k$ is a diagonal matrix, as such the axis-aligned multivariate normal pdf simplifies to the marginal product of Gaussians along each of the $D$ axes as follows,

\begin{equation}
	\begin{aligned}
		\mathcal{N} (X_i, \mu_{k}, \Sigma_k) &=  \prod_{d=1}^D \frac{1}{\sigma_{k,d}\sqrt{2 \pi}} exp \Big( \frac{Z_{i,d} - \mu_{k,d}} {\sigma_{k,d}} \Big)^2 \\[10pt]
		&\text{where} \quad \sigma^2_{k,d} = \Sigma_{k,d,d}
	\end{aligned}
\end{equation}

As there is one cluster per label category, the joint probability for sample $i$ with label assignment $k$, $p(Y_{i,k},X_i)$ is given by the normal pdf of the $k^{th}$ cluster,

\begin{equation}
	\label{eq_pyx}
	p(Y_{i,k},X_i) = \mathcal{N} (Z_i, \mu_{k}, \Sigma_k) \end{equation}

By Bayesian identity, the conditional probability $\hat{Y}_{i,k}=p(Y_{i,k}|X_i)$ can therefore be inferred from Eq. \ref{eq_px} and \ref{eq_pyx} as follows,

\begin{equation}
	\hat{Y}_{i,k} = p(Y_{i,k}|X_i) = \frac{p(Y_{i,k}, X_i)}{p(X_i)}
	\label{eq_conditional}
\end{equation}

The AAGMM layer is implemented as a normal PyTorch \cite{pytorch} module.
It has two parameters updated by backprop.
(1) the explicit cluster centers, a matrix $K \times D$ initialized randomly, and
(2) the diagonal elements of the $D \times D$ matrix $\Sigma_k$ are randomly initialized in the range $[0.9, 1.1]$, which contains the diagonal elements of the GMM Sigma matrix for each cluster.

\subsubsection{KMeans Layer}

We also implement a KMeans final layer which is a more restrictive form of the AAGMM layer.
The KMeans layer is additionally constrained such that the Gaussian covariance matrix $\Sigma_k$ for each cluster center $k$ is the $[D \times D]$ identity matrix. 
This constraint yields spherical cluster centers; similar to how the traditional KMeans algorithm also assumes spherical clusters.
See the published codebase for implementation details about the AAGMM and KMeans layers.

\subsubsection{Relation between K-means, AAGMM and Softmax layers}

\begin{figure}[h]
	\includegraphics[width=8cm]{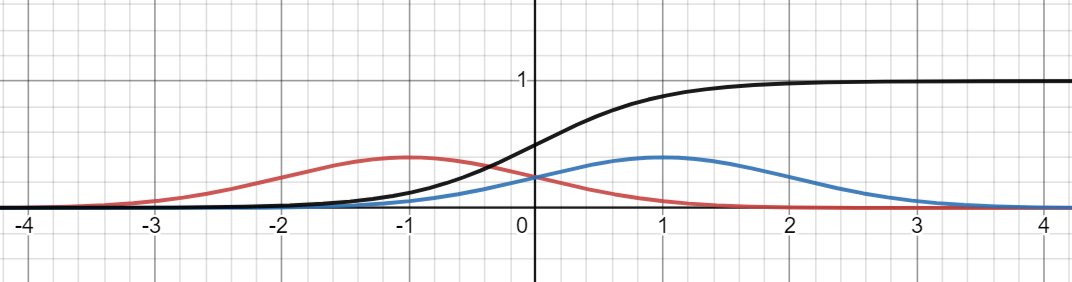}
	\caption{Illustrated relationship between K-means and Softmax Layers for 1D binary classification.  If the joint distributions (blue curve and red curve) follow equivariate normal probability densities, then the conditional distribution (black curve) is softmax.}
	\label{gauss}
\end{figure}

The AAGMM, KMeans and Softmax layers are theoretically related because Softmax is the conditional distribution $p(Y|X)$ that arises when the joint distributions $p(Y,X)$ follow the equivariate normal distributions as modeled by KMeans.  Figure \ref{gauss} shows a simple example of this relationship in one dimension with two labels $A$ (blue curve) and $B$ (red curve), with joint distributions as follows,

\begin{equation}
	\begin{aligned}
		p(Y=A,X) =& \frac{1}{2} \mathcal{N}(X,\mu_A,\sigma) \\
		p(Y=B,C) =& \frac{1}{2} \mathcal{N}(C,\mu_B,\sigma) \\
	\end{aligned}
\end{equation}

In this case the conditional distribution can be described by sigmoid which is a special case of Softmax,

\begin{equation}
	\begin{aligned}
		p(Y=A\; |\;X) \;\; &= \; \textit{sigmoid}(mX + b) \\[6pt]
		&\text{where} \\
		m = \frac{\mu_A - \mu_B}{\sigma^2} \quad &\text{and}  \quad
		b = \frac{\mu_A^2 + \mu_B^2}{2 \sigma^2}
	\end{aligned}
\end{equation}

The AAGMM layer is a generalization of the KMeans layer to allow different diagonal covariance matrices for each cluster.  This gives the AAGMM somewhat greater expressive power than Softmax and KMeans as it can model the joint distribution of latent spaces with different cluster sizes and non-spherical shapes.  

\subsection{Method of Moments Embedding Constraints}
\label{sec:mom}

We introduce and evaluate a series of embedding constraints based on the Method of Moments (MoM) \cite{pearson1936method}.  
For each sample $i$ and each cluster $k$, the joint $p(Y_{i,k},X_i)$ is calculated as in Eq. \ref{eq_pyx} and then used to infer the prior $p(X_i)$ and the conditional $p(Y_{i,k}|X_i)$.
As usual, the conditional probability is trained using cross entropy loss.  
When embedding constraints are omitted, it is possible for the model to learn an accurate decision boundary for the conditional probability without modeling the latent joint distribution.
MoM solves these problems and is an appropriate strategy for semi-parametric models.

The MoM relies on the use of \textit{consistent estimators}, which asymptotically share sample and population statistics.
Assume that $z$ is a finite sample of $n$ elements drawn from infinite population $Z$, then a series of $P$ well-behaved sample statistics $g_p$ should very closely approximate their $P$ population statistic as follows,

\begin{equation}
	\forall p=1 \dots P \quad
	\frac{1}{n} \sum_{i=1}^n g_p(z_i) \approx E(g_p(Z))
\end{equation}

We can therefore constrain the latent representation of our model to approximate a multivariate standard normal distribution.

The centralized moments are a classical choice for the consistent estimator $g_p$ representing the terms of a power series around the mean $\mu$

\begin{equation}
	g_p(Z) = (Z - \mu)^p
\end{equation}

In the univariate standard normal case, the $p^{th}$ order centralized moment constraint is the following.

\begin{equation}
	E\left[ (Z-\mu)^p \right] = 
	\begin{cases} 
		0 &  \text{if} \; p \; \text{is odd} \\
		\sigma^p(p - 1)!! & \text{if} \; p \; \text{is even}
	\end{cases}
\end{equation}

Where '$!!$' represents the double factorial operator.  
By this formula, the univariate unit Gaussian has mean $0$, standard deviation $1$, skew $0$, and kurtosis $3$.

The multivariate standard normal distribution is the marginal product of the univariate standard normal distributions.  As such, if we redefine $Z$, $\mu$, and $p$ to be all $D$ dimensional, then the centralized marginal product moment can be defined as follows,

\begin{equation}
	E\left[g_p(Z - \mu)\right] = E\left[ \prod_{d=1}^D (Z_d - \mu_d)^{p_d} \right]
\end{equation}

Due to independence of the axes, this multivariate population moment can be represented as a product of univariate moments of the individual standard normal distributions as follows,

\begin{equation}
	E\left[ \prod_{d=1}^D (Z_d - \mu_d)^p_d \right] = \prod_{d=1}^D E\left[ (Z_d - \mu_d)^{p_d} \right]
\end{equation}

The error (loss) term associated with the embedding constraint for any moment $p$ is equal to the L2 distance between the sample and population statistics as follows,

\begin{equation}
	\varepsilon_p = \left( \frac{1}{n} \sum_{i=1}^n g_p(z_i) - E(g_p(Z)) \right)^2
\end{equation}

Some moments are more important than others, and must be weighted more heavily.  
First order moments are simply the sample mean, and should be given the greatest weight as an embedding constraint. 
The second order moments form a sample covariance matrix, which ideally should be equal to the identity matrix, but the diagonal terms should be given greater weight than the off-diagonal terms.  
This is because, in a $D \times D$ covariance matrix, there are $D(D-1)$ off diagonal terms, but only $D$, diagonal terms.  
The $p^{th}$ order sample moments form a $p-1$ dimensional hyper-covariance matrix, with terms residing on the intersection of anywhere between $0$ and $p-1$ hyper-diagonals.  
To prevent over-representation of off-diagonal terms and encourage representation of on-diagonal terms, the loss function we use for any given moment term is inversely proportional to the number of moment terms that share the same number of hyper-diagonals.  
This heuristic weighting scheme ensures that the overall contribution of each moment order is not overly influenced by the off-diagonal terms, and that the error weighting is therefore diagonally dominant.
This weighting scheme supports using 0 to 4th order MoM constraints seamlessly and is not a hyper-parameter we expect to require tuning.

\subsection{Manhabalobis Outlier Removal}

The AAGMM layer allows us to detect and remove outliers based on Mahalanobis distance in the latent feature space.  By \textit{outlier}, we are referring to the problem that the pseudolabel learner (i.e. FlexMatch) is simply not yet ready to learn a given unlabeled sample, because the model has only attempted to learn the distribution of labeled and previously pseudolabeled samples up until that point.  Due to small labeled sample size, the labeled and pseudolabeled samples do not fully represent the distribution of the unlabeled samples in early iterations.  Thus, unlabeled samples far from the learned distribution are considered to be ouliers in a given iteration.

In order to implement \textit{outlier removal}, in the context of pseudolabeling, we exclude the unlabeled \textit{detected outlier} sample from gradient updates for any the given iteration of the semi-supervised procedure.  For FlexMatch in particular, all unlabeled samples are augmented with weak and strong RandAugment.  As such we use the weakly augmented samples as input to outlier detection.

Mathematically, we consider an unlabeled sample $x$ to be an outlier if it is distance from all cluster centers $\mu_1 \dots \mu_K$ in terms of Mahalanobis distance based on the cluster covariances $\Sigma_1 \dots \Sigma_K$.  Mahalanobis distance of a given point $x$ to cluster $k$ is defined as follows,

\begin{equation}
	d_{M,k}(x) = \sqrt{ \left(x - \mu_k\right)^T \Sigma_k^{-1} \left(x - \mu_k\right)  }
\end{equation}

Define $X_L$ as the labeled population and $X_U$ as the unlabeled population, with $x \in X_U$ as an unlabeled sample, and $P_{90}$ as the 90th percentile. 
As such, we detect outliers as follows,
\begin{equation}
	\begin{aligned}
		x \in X_U \;\; & \text{is an outlier iff.} \\\
		\textit{max}_k & \; d_{M,k}(x) > \tau \\
		\text{where} \;\; & \tau = P_{90}\left(\textit{max}_k \; d_{M,k}\left(X_L\right)\right)
	\end{aligned}
\end{equation}

\section{Experiments}

We evaluate both AAGMM and KMeans linear layer replacements and the embedding space constraints using our modified FlexMatch \cite{zhang2021flexmatch} on the common SSL benchmarks CIFAR-10 \cite{cifar10} at 40 labels (4 labels per class) and STL-10 \cite{coates2011analysis} at 40 labels. 
We randomly selected 5 seeds a priori for evaluation.
For each algorithm configuration tested, one model was trained per seed.
During each run, the required number of labeled samples are drawn without replacement from the training population of the dataset in a deterministic manner (reproducible with the same seed).
All data not in this labeled subset are used as unlabeled data (i.e., the labels are discarded).

\begin{table*}[ht!]
	\begin{tabularx}{\textwidth}{c|c|XXXX}
		\multicolumn{4}{c}{Mean Test Accuracy Per Method and Dataset} \\ \hline\hline
		Last Layer &   Emb Dim   & \multicolumn{3}{c}{CIFAR-10 at 40 Labels (5 trials)}            \\ 
		\hline
		Embedding Constraint  &  & None & $1st$ Order & $2nd$ Order\\ 
		\hline
		Linear (unmodified FlexMatch) & 128  & $95.03$ \scriptsize{$\pm 0.06$}   &  &  \\
		Linear (FlexMatch w/ Reduced Emb Dim) & 8  & $90.89$ \scriptsize{$\pm 3.24$}      &  & \\
		\hline
		AAGMM & 8  & $94.98$ \scriptsize{$\pm 0.07$}    & $94.64$ \scriptsize{$\pm 0.27$} & $93.58$ \scriptsize{$\pm 2.74$}  \\
		\hline
		KMeans & 8  & $90.15$ \scriptsize{$\pm 6.40$}    & $93.50$ \scriptsize{$\pm 0.62$} & $93.44$ \scriptsize{$\pm 0.50$}   \\
		
		\hline\hline
		Last Layer  &   Emb Dim  & \multicolumn{3}{c}{STL-10 at 40 Labels (3 trials)}            \\ 
		\hline
		\multicolumn{1}{c|}{Embedding Constraint} &  & None & $1st$ Order & $2nd$ Order  \\ 
		\hline
		Linear (unmodified FlexMatch) & 128  & $70.85$ \scriptsize{$\pm 4.16$}   &  &   \\
		\hline
		AAGMM & 8  & $58.79$ \scriptsize{$\pm 11.00$}    & $71.11$ \scriptsize{$\pm 7.60$} & $70.40$ \scriptsize{$\pm 6.39$}  \\
	\end{tabularx}
	\caption{Mean test accuracy \% for CIFAR-10 and STL-10 SSL benchmarks comparing various configurations of our method. The FlexMatch results in the table is drawn from the publication. For CIFAR-10, the WideResNet model used by FlexMatch has an embedding size of 128 dimension. Results for a given order of embedding constraint include all lower constraints.}
	\label{table1}
\end{table*}

\subsection{Hyper-Parameters}
\label{hyperparams}

All CIFAR-10 models were trained with the standard benchmark WideResNet28-2 architecture \cite{zagoruyko2016wide}.
All STL-10 models were trained with the standard benchmark WideResNet37-2 architecture \cite{zagoruyko2016wide}.
This work leverages the published FlexMatch \cite{zhang2021flexmatch} code, hyper-parameters, and training configurations within the USB Framework \cite{wang2022usb}.
However, due to the higher training instability of the AAGMM layers compared to a linear layer, the gradient norm was clipped to 1.0.
Despite specific attention to computing the AAGMM and embedding constraints in a numerically stable manner, they are still less stable during backprop than a simple linear layer.  
Gradient clipping was especially important when there were latent points that were multiple standard deviations away from the cluster centers.  
In this case, the gradient of the Gaussian probability density function converges rapidly toward zero, which can affect the division step of equation \ref{eq_conditional}.  
We have almost entirely overcome this issue in the code by using laws of exponents in order to normalize the denominator to be greater or equal to 1 prior to division, but in extreme cases of latent points distant from cluster centers, the gradient clipping is still necessary to achieve stable gradient descent.

We believe that the rapid decrease in slope of the Gaussian distribution pdf for points that are multiple standard deviations away from the mean.  
As the conditional distribution involves calculating a division step, it is very likely that such a division may become less stable for points that are not near the mean.  

This work includes an exploration of how various latent embedding dimensionalities affect the generative linear layer replacement.
As such, the model architecture was modified with a single additional linear layer before the output to project the baseline model embedding dimension (128 for WideResNes28-2) down to a reduced 8 dimensional space.
Results listed with an embedding dimensionality of 128 do not include the additional linear layer which reduces the latent dimensionality. 

Due to exponential GPU memory requirements with each successive MoM moment, only the 8D embedding can operate with higher order MoM embedding constraints. 
While our method can place constraints on any number of moments, we only explored MoM constraints up to the second order in this paper. 
Results for any given order of embedding constraint include all lower constraints.

\subsection{CIFAR-10 and STL-10 Results}

While current SOTA on CIFAR-10 at 250 labels is close to fully supervised accuracy, the 40 label case provides a far more challenging task. 
Table \ref{table1} summarizes the relative performance of our various configurations.
As we simply extended FlexMatch, our hyper-parameter selection is likely sub-optimal, and further experimentation might yield improvements.
The 'Linear (unmodified FlexMatch)' row in Table \ref{table1} represents the baseline fully connected linear last layer exactly as FlexMatch published. 
The 'Linear (FlexMatch w/ Reduced Emb Dim)' row in Table \ref{table1} represents FlexMatch performance where its final Linear layer produced a latent embedding dimensionality of 8 (instead of the stock model latent dimensionality of 128).
Its worth noting that the significantly smaller embedding dimensionality reduced average model accuracy by $5\%$.
That accuracy is restored by replacing the Linear layer with an AAGMM layer; despite keeping the reduced latent embedding dimensionality.
That trend is mirrored with STL-10, where the $70.85\%$ FlexMatch accuracy drops to $58.79\%$ using a latent embedding dimensionality of 8 (and the unconstrained AAGMM layer), which the 1st order embedding constrained version of AAGMM restores to $71.11\%$ accuracy.

We see that the AAGMM final layer consistently out performs the KMeans final layer for the CIFAR-10 Test Accuracy with 40 labels.  
We furthermore see that the KMeans layer performs significantly better with the 1st and 2nd order MoM constraint, as compared with no constraints.  

For STL-10 the AAGMM final layer ($71.11\%$) improved upon the FlexMatch \cite{zhang2021flexmatch} result ($70.85\%$), though within the margin of error.
Additionally, both 1st and 2nd order MoM constraints significantly improved the upon the AAGMM with no embedding constraints.

The modeled cluster centers vary in quality between individual model runs of the AAGMM layer due to the stochasticity of the training process.
Figure \ref{fig:cifar10tsne} (a \& b) showcases degenerate cluster centers.
The Figure \ref{fig:cifar10tsne} (c \& d) AAGMM model learned cluster centers that are an adequate approximation of the underlying data, where the embedding constraints encourage cluster centers which are better aligned with the underlying data.
It is worth noting that we did not observe the KMeans layer learning non-degenerate cluster centers without an embedding constraint.
In contrast, the AAGMM layer can, under some circumstances, learn viable cluster centers.
To quantify the modeled cluster compactness, we measure the average L2 distance from each test data point to its assigned cluster as shown in Table \ref{compactness}.

\begin{table}[h!]
	\begin{tabular}{c|c|c|c|c}
		
		\multicolumn{5}{c}{Latent Embedding Space Cluster Compactness} \\
		\hline\hline
		Dataset & Last Layer & None & 1st MoM & 2nd MoM \\
		\hline
		CIFAR-10 & AAGMM & $1.03$ & $0.58$ & $0.53$ \\
		CIFAR-10 & KMeans & $18.41$ & $0.96$ & $1.06$\\
		\hline
		STL-10 & AAGMM & $2.32$ & $0.79$ & $0.76$ \\
	\end{tabular}
	\caption{Average L2 distance from each test data point to its assigned cluster center.}
	\label{compactness}
\end{table}

To place our results in context with the current SSL SOTA for CIFAR-10 and STL-10 at 40 labels Table \ref{sslcifar10} compares against the current best methods, demonstrating that this methodology is nearly competitive with for CIFAR-10, but still requires improvement and tuning for STL-10. 

\begin{table}[htbp]
	\begin{tabular}{c|cc}
		& CIFAR-10  &STL-10 \\ 
		Method & (40 Labels) & (40 Labels) \\
		\hline
		\hline
		FixMatch\cite{sohn2020fixmatch}   & $13.81$ \scriptsize{$\pm3.37$}   & $35.97$ \scriptsize{$\pm4.14$}     \\
		FlexMatch\cite{zhang2021flexmatch}  & $4.97$ \scriptsize{$\pm0.06$}    & $29.15$ \scriptsize{$\pm4.16$}    \\
		FreeMatch\cite{wang2022freematch}  & $4.90$ \scriptsize{$\pm0.29$}    & $15.56$ \scriptsize{$\pm0.55$}    \\
		SimMatchV2\cite{zheng2023simmatchv2} & $4.90$ \scriptsize{$\pm0.04$}    & $15.85$ \scriptsize{$\pm2.62$}    \\ \hline
		\textbf{AAGMM+None}    & $5.02$ \scriptsize{$\pm 0.07$}           & $41.21$ \scriptsize{$\pm11.00$}  \\
		\textbf{AAGMM+1stOrder}   & $5.36$ \scriptsize{$\pm 0.27$}           & $28.89$ \scriptsize{$\pm .60$}  
		
	\end{tabular}
	\caption{Error rate \% for CIFAR-10 and STL-10 SSL benchmarks with 40 labels, comparing to state of the art results. Results for previously published methods are drawn from USB \cite{wang2022usb} except for FreeMatch \cite{wang2022freematch} and SimMatchV2 \cite{zheng2023simmatchv2} publications.}
	\label{sslcifar10}
\end{table}

\section{Discussion}

Although our preliminary results with the proposed AAGMM and MoM achieve high accuracy relative to SOTA, this was achieved without Mahalanobis outlier detection as documented in Table \ref{outlierRemoval}. 
When the outlier detection was enabled, we believe the reduced accuracy is due to: 1. the 90th percentile distance threshold being too aggressive and filtering too much signal relative to noise, and 2. the need for an adaptive outlier detection threshold. 
In early epochs, an aggressive outlier detection threshold is viable because the model will not adequately fit many of the unlabeled samples, but as the model converges the fit improves reducing the need for outlier removal.  As such, we believe that the aggressive outlier filtering is removing too many inliers, particularly in later epochs.

\begin{table}[ht!]
	\begin{tabular}{c|c|c}
		
		\multicolumn{3}{c}{AAGMM (8D) Mean Test Accuracy With Outlier Removal} \\ \hline\hline
		\multicolumn{3}{c}{CIFAR-10}            \\ 
		\hline
		Outlier Threshold & None & 90th Percentile \\
		\hline
		MoM: None & $94.98$ \scriptsize{$\pm 0.07$} & $94.9$ \scriptsize{$\pm 0.125$} \\
		MoM: $1st$ Order & $94.64$ \scriptsize{$\pm 0.27$} & $87.70$ \scriptsize{$\pm 2.96$}\\
		MoM: $2nd$ Order & $93.58$ \scriptsize{$\pm 2.74$} & $87.25$ \scriptsize{$\pm 2.51$} \\
		
		\hline
		\hline
		\multicolumn{3}{c}{STL-10}            \\ 
		\hline
		Outlier Threshold & None & 90th Percentile \\
		\hline
		MoM: None & $58.79$ \scriptsize{$\pm 11.00$} & $57.50$ \scriptsize{$\pm 12.75$} \\
		MoM: $1st$ Order & $71.11$ \scriptsize{$\pm 7.60$} & $64.18$ \scriptsize{$\pm 3.82$}\\
		MoM: $2nd$ Order & $70.40$ \scriptsize{$\pm 6.39$} & $65.90$ \scriptsize{$\pm 4.09$} \\
	\end{tabular}
	\caption{Mean test accuracy \% for CIFAR-10 and STL-10 SSL benchmarks comparing an 8D embedding AAGMM final layer with and without outlier removal during training. Results for a given order of embedding constraint include all lower constraints.}
	\label{outlierRemoval}
\end{table}

The proposed MoM embedding constraint has at least one significant downside, by requiring exponentially increasing amounts of GPU memory for each successive moment penalty included as shown in Table \ref{gpuRam}.

\begin{table}[h!]
	\begin{tabular}{c|c|c|c|c|c}
		
		\multicolumn{6}{c}{AAGMM Training GPU Memory Requirements (in GiB)} \\
		\hline\hline
		Emb Dim & None & 1st & 2nd & 3rd & 4th \\
		\hline
		$8$ & $7.72$ & $7.71$ & $7.70$ & $7.76$ & $8.76$ \\
		$32$ & $7.71$ & $7.71$ & $7.79$ & $13.15$ & $>20.47$ \\
		
	\end{tabular}
	\caption{GPU RAM utilization in GiB on Nvidia RTX A4500 with 20.47GiB of VRAM evaluated on CIFAR10 using WRN28-2 with a batch size of 64. Each row shows the results for a given embedding dimensionality, and each row a MoM embedding constraint order, where "None" indicates a stock linear layer.}
	\label{gpuRam}
\end{table}

This limits the current practicality of these MoM constraints. 
Additional optimization and/or avoiding the explicit creation of both the $n^{th}$ order moment and its target value on GPUs would likely improve the usability.

Semi-supervised learning is highly sensitive to both which samples are selected form the labeled population \cite{sohn2020fixmatch} and the stochasticity of the training process itself.

Future work in this area will explore alternate outlier removal strategies, including thresholding the unlabeled samples based on their latent sample probability $p(X)$ as opposed to latent Mahalanobis distance.  
The Mahalanobis distance is part of the exponential term in the calculation of the multivariate Gaussian PDF. 
As such, we expect that a removing outliers based on low $p(X)$ is likely to perform comparably to removing samples based on far Mahalanobis distance, in the special case that all clusters share a similar determinant $det\left(\Sigma_k \right)$.  
Additionally, we plan to explore how to best take advantage of the better behaved latent embedding space to improve data efficiency for model training. 

We demonstrate a novel fully differentiable Axis-Aligned Gaussian Mixture Model with Method of Moments based latent embedding space constraints can be applied to semi-supervised learning.  
The combination of these techniques enables outlier detection strategies that would otherwise not be possible with a traditional softmax discriminator approach.
This preliminary work constructs these novel layers with the associated constraints and demonstrates reasonable performance on challenging benchmark semi-supervised learning tasks, while opening the door for future outlier detection strategies that can make semi-supervised learning more robust to large and diverse unlabeled sample distributions.

\textbf{Acknowledgements.} We would like to thank the Frost Institute for Data Science and Computing for their support.

{
	\small
	\bibliographystyle{ieeenat_fullname}
	\bibliography{refs}
}

\end{document}